\def\year{2022}\relax
   \documentclass[letterpaper]{article}
   \usepackage{aaai22} 
   \usepackage{times} 
   \usepackage{helvet} 
   \usepackage{courier} 
   \usepackage[hyphens]{url} 
   \usepackage{graphicx} 
   \usepackage{graphicx}  
   \usepackage{natbib}  
   \usepackage{caption}  
   \DeclareCaptionStyle{ruled}%
     {labelfont=normalfont,labelsep=colon,strut=off}
\usepackage{bm}
\usepackage{physics}
\usepackage{mathtools}
\usepackage{amsfonts}  
\usepackage{bbm}
\mathtoolsset{showonlyrefs}
\usepackage{multirow}
\usepackage{array}
\newcolumntype{C}[1]{>{\centering\arraybackslash$}p{#1}<{$}}
\usepackage{algorithmic}
\usepackage[vlined,ruled,linesnumbered]{algorithm2e}
\usepackage{subfigure}
\usepackage{xcolor}
\usepackage{comment}

\definecolor{mygreen}{rgb}{0,.6,0}
\definecolor{myyellow}{rgb}{1,.7,0}

\newcommand{\KL}[1]{{\color{red}KL: {#1}}}
\newcommand{\Red}[1]{{\color{red}{#1}}}
\newcommand{\Blue}[1]{{\color{blue}{#1}}}
\newcommand{\Green}[1]{{\color{mygreen}{#1}}}
\newcommand{\Yellow}[1]{{\color{myyellow}{#1}}}

\newcommand{\Transpose}{^{\mathsf{T}}}

\newcommand{\StateSymb}{x}
\newcommand{\State}{\bm{\StateSymb}}
\newcommand{\StateMat}{X}

\newcommand{\HiddenStateSymb}{z}
\newcommand{\HiddenState}{\bm{\HiddenStateSymb}}

\newcommand{\Dictionary}{\Phi}
\newcommand{\DictionarySymbol}{\phi}
\newcommand{\DictionaryVec}{\pmb{\DictionarySymbol}}

\newcommand{\PolyTwo}{\mathbbm{P}_2}

\newcommand{\Coeffs}{\xi}
\newcommand{\CoeffsMat}{\Xi}

\newcommand{\DimState}{n}
\newcommand{\DimTime}{m}
\newcommand{\DimPoly}{p}
\newcommand{\Degree}{d} 

\newcommand{\nTrain}{n_{\mathrm{train}}}

\newcommand{\nBatch}{n_{\mathrm{batch}}}

\newcommand{\TimeSymb}{t}
\newcommand{\TimeStep}{\Delta t}

\newcommand{\Velocity}{\bm{f}}
\newcommand{\VelocityNN}{\Velocity_{\NNParams}}

\newcommand{\NNParams}{\Theta}

\newcommand{\Threshold}{\tau}

\newcommand{\Hamiltonian}{\mathcal H}

\newcommand{\PoissonBracket}[2]{\{#1,#2\}}
\newcommand{\PoissonBracketBi}[2]{\pdv{#1}{\State}\PoissonMatrix\pdv{#2}{\State}}

\newcommand{\PoissonMatrix}{L}

\newcommand{\IrrBracket}[2]{[#1,#2]}
\newcommand{\IrrBracketBi}[2]{\pdv{#1}{\State}\FrictionMatrix\pdv{#2}{\State}}
\newcommand{\FrictionMatrix}{M}

\newcommand{\TotalEnergy}{E}

\newcommand{\TotalEnergyNN}{\TotalEnergy_{\theta_1}}

\newcommand{\Entropy}{S}

\newcommand{\Penalty}{\lambda}

\newcommand{\maxepoch}{n_{\max}}

\newcommand{\DataSet}{\mathcal D}
\newcommand{\TrainSet}{\DataSet_{\mathrm{train}}}

\newcommand{\LossSymb}{\mathcal L}

\newcommand{\LBatch}{\ell_{\mathrm{batch}}}

\newcommand{\PolySet}{\mathcal P}

\begin{document}
%
\title{Structure-preserving Sparse Identification of Nonlinear Dynamics\\ for Data-driven Modeling}
\author{Kookjin Lee\textsuperscript{\rm 1}, Nathaniel Trask\textsuperscript{\rm 2}, and Panos Stinis\textsuperscript{\rm 3}}
\affiliations{
\textsuperscript{\rm 1} School of Computing and Augmented Intelligence, Arizona State University, Tempe, AZ 85281\\
\textsuperscript{\rm 2} Center for Computing Research, Sandia National Laboratories, Albuquerque, NM 87123\\
\textsuperscript{\rm 3} Pacific Northwest National Laboratory, Richland, WA 99354\\
kookjin.lee@asu.edu, natrask@sandia.gov, panagiotis.stinis@pnnl.gov
}
\maketitle
\begin{abstract}
Discovery of dynamical systems from data forms the foundation for data-driven modeling and recently, structure-preserving geometric perspectives have been shown to provide improved forecasting, stability, and physical realizability guarantees. We present here a unification of the Sparse Identification of Nonlinear Dynamics (SINDy) formalism with neural ordinary differential equations. The resulting framework allows learning of both "black-box" dynamics and learning of structure preserving bracket formalisms for both reversible and irreversible dynamics. We present a suite of benchmarks demonstrating effectiveness and structure preservation, including for chaotic systems.
\end{abstract}

\section{Introduction}

A number of scientific machine learning (ML) tasks seek to discover a dynamical system whose solution is consistent with data (e.g. constitutive modeling \cite{patel2020thermodynamically,karapiperis2021data,ghnatios2019data,masi2021thermodynamics}, reduced-order modeling \cite{chen2021physics,lee2021deep,wan2018data}, physics-informed machine learning \cite{karniadakis2021physics,wu2018physics}, and surrogates for performing optimal control \cite{alexopoulos2020digital}). A major challenge for this class of problems is the preservation of both numerical stability and physical realizability when performing out of distribution inference (i.e. extrapolation/forecasting). Unlike traditional ML for e.g. image/language processing tasks, engineering and science models pose strict requirements on physical quantities to guarantee properties such as conservation, thermodynamic consistency and well-posedness of resulting models \cite{baker2019workshop}. These structural constraints translate to desirable mathematical properties for simulation, such as improved numerical stability and accuracy, particularly for chaotic systems \cite{lee2021machine,trask2020enforcing}.

While so-called physics-informed ML (PIML) approaches seek to impose these properties by imposing soft physics constraints into the ML process, many applications require structure preservation to hold exactly; PIML requires empirical tuning of weighting parameters and physics properties hold only to within optimization error, which typically may be large \cite{wang2020understanding,rohrhofer2021pareto}. Structure-preserving machine learning has emerged as a means of designing architectures such that physics constraints hold exactly by construction \cite{lee2021machine,trask2020enforcing}. By parameterizing relevant geometric or topological structures, researchers obtain more data-efficient hybrid physics/ML architectures with guaranteed mathematical properties.

In this work we consider geometric structure preservation for dynamical systems \cite{hairer2006geometric,marsden1995introduction}. Reversible bracket formalisms (e.g. Hamiltonian/Lagrangian mechanics) have been shown effective for learning reversible dynamics \cite{toth2019hamiltonian,cranmer2020lagrangian,lutter2018deep,chen2019symplectic,jin2020sympnets,tong2021symplectic,zhong2021benchmarking,chen2021data,bertalan2019learning}, while dissipative metric bracket extensions provide generalizations to irreversible dynamics in the metriplectic formalism \cite{lee2021machine,desai2021port,hernandez2021structure,yu2020onsagernet,zhong2020dissipative}. Accounting for dissipation in this manner provides important thermodynamic consistency guarantees: namely, discrete versions of the first and second law of thermodynamics along with a fluctuation-dissipation theorem for thermal systems at microscales. Training these models is however a challenge and leads to discovery of non-interpretable \textit{Casimirs} (generalized entropy and energy functionals describing the system). 

For $\State(\TimeSymb) \in \mathbb{R}^{\DimState}$ denoting the state of a system at time $\TimeSymb$, $\Velocity$ denoting a velocity with potentially exploitable mathematical structure, and $\NNParams$ denoting trainable parameters, we consider learning of dynamics of the form
\begin{equation}
    \dv{\State(\TimeSymb)}{\TimeSymb} = \VelocityNN(\State).
\end{equation}
We synthesize the Sparse Identification of Nonlinear Dynamics (SINDy) formalism \cite{brunton2016discovering} with neural ordinary differential equations (NODEs) \cite{chen2018neural} to obtain a framework for learning dynamics. This sparse dictionary-learning formalism allows learning of either interpretable $\Velocity$ when learning "black-box" ODEs, or for learning more complicated bracket dynamics for $\Velocity$ which describe structure-preserving reversible and irreversible systems. Our main technical contributions include: 
\begin{itemize}
    \item a novel extension of SINDy into neural ODE settings, including a training strategy with L1 weight decay and pruning,
    \item the first application of SINDy with structure-preservation, including: black-box, Hamiltonian, GENERIC, and port-Hamiltonian formulations
    \item empirical demonstration on the effectiveness of the proposed algorithm for a wide array of dynamics spanning: reversible, irreversible, and chaotic systems.
\end{itemize}

\section{Sparse Identification of Nonlinear Dynamics (SINDy)}
The Sparse Identification of Nonlinear Dynamics (SINDy) method aims to identify the dynamics of interest using a sparse set of dictionaries such that 
\begin{equation}
    \VelocityNN(\State) = (\DictionaryVec(\State)\Transpose \CoeffsMat)\Transpose,
\end{equation}
where $\CoeffsMat \in \mathbb{R}^{\DimPoly\times\DimState}$ is the coefficient matrix and $\DictionaryVec(\HiddenState) \in \mathbb{R}^{1\times\DimPoly}$ denotes a ``library'' vector consisting of candidate functions, e.g., constant, polynomials, and so on. 

From measurements of states (and derivatives), we can construct a linear system of equations:
\begin{equation}
    \dot{\StateMat} = \Dictionary(\StateMat) \CoeffsMat
\end{equation}
where $X \in \mathbb{R}^{\DimTime\times\DimState}$ and $\dot{\StateMat}\in \mathbb{R}^{\DimTime\times\DimState}$ are collections of the states $\State(\TimeSymb)$ and the (numerically approximated) derivatives $\dv{\State}{\TimeSymb}$ sampled at time indices $\{\TimeSymb_1,\ldots,\TimeSymb_{\DimTime}\}$ such that 
\begin{equation}
    [\StateMat]_{ij} = \StateSymb_j(\TimeSymb_i), \quad \text{and} \quad [\dot{\StateMat}]_{ij} = \dv{\StateSymb_{j}}{\TimeSymb} {}(\TimeSymb_i).
\end{equation}
A library, $\Dictionary(\StateMat)\in\mathbb{R}^{\DimTime\times\DimPoly}$, consists of candidate nonlinear functions, e.g., constant, polynomial, and trigonometric terms: 
\begin{equation}
    \Dictionary(\StateMat) = \begin{bmatrix} 
    \mathbbm{1} & \StateMat & \PolyTwo(\StateMat) & \cdots & \cos(\StateMat) & \cdots 
    \end{bmatrix},
\end{equation}
where $\PolyTwo$ is a function of the quadratic nonlinearities such that the $i$th row of $\PolyTwo (\StateMat)$ is defined as, for example, with $\DimState=3$, 
\begin{equation}
    \begin{split}
    [\PolyTwo (\StateMat)]\Transpose_{i} = [\StateSymb_1^2&(\TimeSymb_i), \StateSymb_1(\TimeSymb_i) \StateSymb_2(\TimeSymb_i), \\&\StateSymb_1(\TimeSymb_i)\StateSymb_3(\TimeSymb_i), \StateSymb_2^2(\TimeSymb_i), \StateSymb_2(\TimeSymb_i)\StateSymb_3(\TimeSymb_i), \StateSymb_3^2(\TimeSymb_i)].
    \end{split}
\end{equation} 
Lastly, $\CoeffsMat\in\mathbb{R}^{\DimPoly\times\DimState}$ denotes the collection of the unknown coefficients $\CoeffsMat = [\Coeffs_1,\ldots,\Coeffs_{\DimState}]$, where $\Coeffs_i\in\mathbb{R}^{\DimPoly}$, $i=1,\ldots,\DimState$ are sparse vectors.

Typically, only $\StateMat$ is available and $\dot{\StateMat}$ is approximated numerically. Thus, in SINDy,  $\dot{\StateMat}$ is considered to be noisy, which leads to a new formulation: \begin{equation}
    \dot{\StateMat} = \Dictionary(\StateMat) \CoeffsMat + \eta Z,
\end{equation}
where $Z$ is a matrix of independent identically distributed unit normal entries and $\eta$ is noise magnitude. To compute the sparse solution $\CoeffsMat$, SINDy employs either the  least absolute shrinkage and selection operator (LASSO) \cite{tibshirani1996regression} or sequential threshold least-squares method \cite{brunton2016discovering}. LASSO is an L1-regularized regression technique and the sequential threshold least-squares method is an iterative algorithm which repetitively zeroes out entries of $\CoeffsMat$ and solves least-squares problems with remaining entries of $\CoeffsMat$.

\section{Neural SINDy}
One evident weakness of SINDy is that it requires the derivatives of the state variable either empirically measured or numerically computed. To resolve this issue, we propose to use the training algorithm introduced in neural ordinary differential equations (NODEs) \cite{chen2018neural} with L1 weight decay. In addition, we propose the magnitude-based pruning strategy to retain only the dictionaries with significant contributions.

\subsection{Neural Ordinary Differential Equations}
NODEs \cite{chen2018neural} are a family of deep neural network models that parameterize the time-continuous dynamics of hidden states $\HiddenState(\TimeSymb)$ using a system of ODEs:
\begin{equation}
    \dv{\HiddenState(\TimeSymb)}{\TimeSymb} = \VelocityNN (\HiddenState,\TimeSymb),
\end{equation}
where $\HiddenState(\TimeSymb)$ is a time-continuous representation of a state, $\VelocityNN$ is a parameterized (trainable) velocity function, and $\NNParams$ is a set of neural network weights and biases.

In the forward pass, given the
initial condition $\HiddenState(0)$, a hidden state at any time index can be obtained by solving the initial value problem (IVP). A black-box time integrator can be employed to compute the hidden states with the desired accuracy:

\begin{equation}
    \tilde{\HiddenState}(\TimeSymb_1), \ldots, \tilde{\HiddenState}(\TimeSymb_{\DimTime}) = \mathrm{ODESolve}(\HiddenState(0), \VelocityNN,\TimeSymb_1, \ldots,\TimeSymb_{\DimTime}).
\end{equation}
The model parameters $\NNParams$ are then updated with an optimizer, which minimizes a loss function measuring the difference between the output and the target variables. 

\paragraph{Dictionary-based parameterization}
As in the original SINDy formulation, we assume that there is a set of candidate nonlinear functions to represent the dynamics  
\begin{equation}\label{eq:dict}
    \VelocityNN(\HiddenState) = (\DictionaryVec(\HiddenState)\Transpose \CoeffsMat)\Transpose,
\end{equation}
where, again, $\CoeffsMat \in \mathbb{R}^{\DimPoly\times\DimState}$ is a trainable coefficient matrix and $\DictionaryVec(\HiddenState) \in \mathbb{R}^{1\times\DimPoly}$ denotes a library vector. In this setting, the learnable parameters of NODE become $\NNParams=\CoeffsMat$. 

\paragraph{Sparsity inducing loss: L1 weight decay (Lasso)}

Following SINDy, to induce sparsity in $\CoeffsMat$ we add an L1 weight decay to the main loss objective for the mean absolute error:
\begin{equation}\label{eq:loss}
    \LossSymb = \frac{1}{\nTrain} \sum_{\ell=1}^{\nTrain} \sum_{i=1}^{\DimTime} \left\| \HiddenState^{(\ell)}(\TimeSymb_i) - \tilde{\HiddenState}^{(\ell)}(\TimeSymb_i) \right\|_1 +  \Penalty \left\| \CoeffsMat \right\|_1,
\end{equation}
where $ \left\| X \right\|_1 = \sum_{l,k} \left| \left[ X \right]_{lk} \right|$.

\paragraph{Pruning}
Taking linear combinations of candidate functions in Eq.~\eqref{eq:dict} admits implementation as a linear layer, specified by $\CoeffsMat$, which does not have biases or nonlinear activation. To further sparsify the coefficient matrix $\CoeffsMat$, we employ the magnitude-based pruning strategy: 
\begin{equation}\label{eq:prune}
    [\CoeffsMat]_{lk} = 0\quad \mathrm{if}\quad [\CoeffsMat]_{lk} < \Threshold
\end{equation}
We find that pruning is essential to find the sparse representation of $\CoeffsMat$ and will provide empirical evidence obtained from the numerical experiments. In next Section, we detail how we use the pruning strategy during the training process. 

\subsection{Training}
We employ mini-batching to train the proposed neural network architecture. For each training step, we randomly sample $\nBatch$ trajectories from the training set and then randomly sample initial points from the selected $\nBatch$ trajectories to assemble $\nBatch$ sub-sequences of length $\LBatch$. We solve IVPs with the sample initial points, measure the loss (Eq.~\eqref{eq:loss}), and update the model parameters $\NNParams$ via Adamax. After the update, we prune the model parameters using the magnitude-based pruning as shown in \eqref{eq:prune}. Algorithm \ref{alg:train} summarizes the training procedure.

\begin{algorithm}[!ht]
\small
\SetAlgoLined
\caption{Neural SINDy training}\label{alg:train}
Initialize $\NNParams$\\
\For{$(i = 0;\ i <\maxepoch ;\ i = i + 1)$}{
    Sample $\nBatch$ trajectories randomly from $\TrainSet$  \\
    Sample initial points randomly from the sampled trajectories: $\State^{r}_{s(r)}$, $s(r)\in[0,\ldots,\DimTime-\LBatch-1]$ for $r=1,\ldots,\nBatch$\\
    $\tilde{\State}(\TimeSymb_1)\!,\ldots,\!\tilde{\State}(\TimeSymb_\DimTime) \!\!=\!\! \mathrm{ODESolve}(\State^{r}_{s(r)}, \VelocityNN,\TimeSymb_1, \!\ldots\!, \TimeSymb_{\DimTime})$, for $r=1,\ldots,\nBatch$\\
    Compute the loss $\LossSymb$ (Eq. \eqref{eq:dict}) \\
    Update $\NNParams$ via SGD\\
    Prune $\NNParams$ based on the magnitude (Eq. \eqref{eq:prune})
    }
\end{algorithm}

\begin{table*}[t]
\centering
\begin{tabular}{l|l|l}
    \hline
    Eq. name & Ground truth & Identified \\
    \hline
    \multirow{2}{*}{Hyperbolic}  & $\dot{x} = -0.05x$ & $\dot{x} = -0.050006x$\\ 
     & $\dot{y} = x^2-y$ & $\dot{y} = 1.000063 x^2-1.000063y$\\
    \hline
    \multirow{2}{*}{Cubic oscillator} & $\dot{x} = -0.1x^3 +2y^3$ & $\dot{x} =  -0.100075x^3 +2.000019y^3$\\
     & $\dot{y} = -2x^3 -0.1y^3$ & $\dot{y} = -1.999906 x^3 -0.099915y^3$ \\
    \hline
    \multirow{2}{*}{Van der Pol} & $\dot{x} = y$ & $\dot{x} = 1.000015 y$\\
     & $\dot{y} = -x+2y-2x^2y$ & $\dot{y} = -1.000043x+2.000271y-1.999807x^2y$\\
     \hline
     \multirow{3}{*}{Hopf bifurcation} & $\dot{\mu}=0$& $\dot{\mu}=0$\\
     & $\dot{x} = \mu x + y - x^3 -xy^2$& $\dot{x} = 0.999237 \mu x + 1.000174 y -0.999523 x^3 -0.999582 xy^2$\\
     & $\dot{y} = \mu y - x - yx^2 -y^3$ & $\dot{y} = 0.999515 \mu y -0.999821 x -0.999276 yx^2 -0.999409y^3$\\
     \hline
     \multirow{3}{*}{Lorenz} & $\dot{x} = -10x+10y$ & $\dot{x} = -10.000108x+9.999878y$ \\
     & $\dot{y} = 28 x - xz - y$ & $\dot{y} = 27.999895 x -0.999998 xz -0.999883 y$ \\
     & $\dot{z} = xy - \frac{8}{3}z$ & $\dot{z} = 1.000017 xy -2.666697 z$ \\
     \hline
\end{tabular}
\caption{[nSINDy experiments] The  equation names, the ground truth equations, and the identified equations by using nSINDy.}
\label{tab:exp_set1}
\end{table*}

\section{Experimental results with neural SINDy}

We implement the proposed method in \textsc{Python} 3.7.2 and \textsc{PyTorch} 1.9.0 \cite{paszke2019pytorch}. In all training, we use Adamax \cite{kingma2015Adam} with an initial learning rate 0.01 and  use exponential learning rate decay with the multiplicative factor, 0.9987. In the following experiments, we set the penalty weight as $\Penalty=10^{-4}$ and the pruning threshold as  $\Threshold=10^{-6}$. For \text{ODESolve}, we use the Dormand--Prince method (dopri5) \cite{dormand1980family} with relative tolerance $10^{-7}$ and absolute tolerance $10^{-9}$ unless otherwise specified. All experiments are performed on \textsc{Macbook pro} with \textsc{M1} CPU and 16 GB memory.

\paragraph{Dictionary construction} 
In the following experiment, we employ a set of polynomials, $\PolySet_{\DimState,\Degree}$, where $\DimState$ is the number of variables and $\Degree$ is the maximal total degree of polynomials in the set. An example set, $\PolySet_{3,2}$ is given as
\begin{equation}
    \PolySet_{3,2} = \{ 1, \StateSymb_1, \StateSymb_1^2, \StateSymb_1 \StateSymb_2, \StateSymb_1 \StateSymb_3, \StateSymb_2, \StateSymb_2^2, \StateSymb_2 \StateSymb_3, \StateSymb_3^2 \}.
\end{equation}

\subsection{Experiment set}
We examine the performance of the proposed method for a number of dynamical systems (see Table \ref{tab:exp_set1} for the equations):
\begin{itemize}
    \item Hyperbolic example,
    \item Cubic oscillator,
    \item Van der Pol oscillator,
    \item Hopf bifurcation,
    \item Lorenz system.
\end{itemize}

To generate data, we base our implementation on the code from \cite{liu2020hierarchical}: generating 1600 training trajectories, 320 validating trajectories, and 320 test trajectories. For the first four example problems, we use time step size $\TimeStep=0.01$ and total simulation time $\TimeSymb\in [0,51.20]$. For Lorenz, we use $\TimeStep=0.0005$ and $\TimeSymb\in [0,2.56]$. For training, we use $\maxepoch=500$ and $\maxepoch=2000$ for the first four example problems and the last example problem, respectively. In Appendix, a comparison against a ``black-box'' feed-forward neural network with fully-connected layers is presented; we report  the performance of both the black-box neural network and the proposed network measured in terms of time-instantaneous mean-squared error (MSE). 

\begin{figure*}[!tb]
    \centering
     \includegraphics[scale=.55]{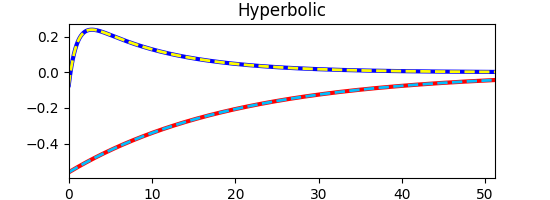}
    \includegraphics[scale=.55]{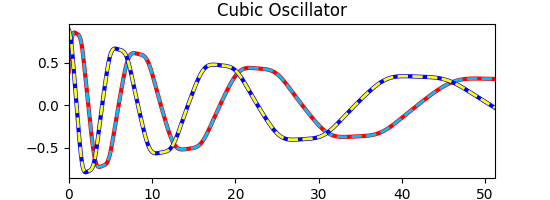}\\
    \includegraphics[scale=.55]{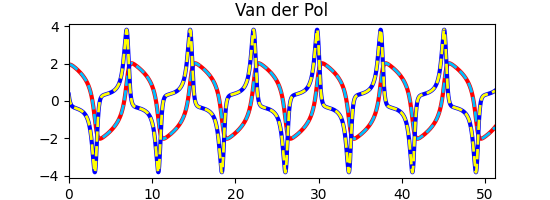}
    \includegraphics[scale=.55]{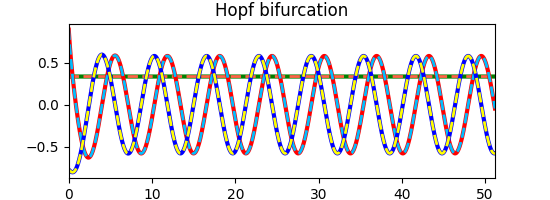}\\
    \includegraphics[scale=.55]{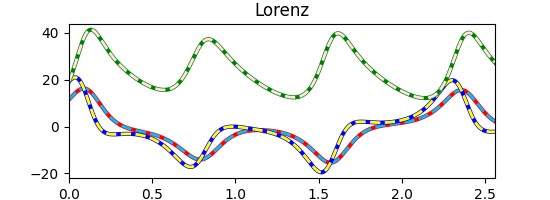}
    \includegraphics[scale=.55]{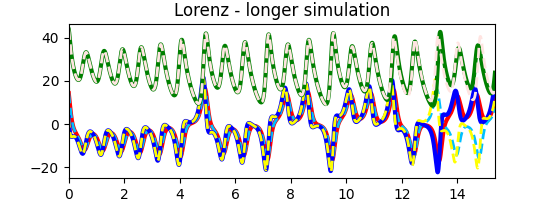}
    \caption{[nSINDy experiments] Examples of reference trajectories (solid lines) and computed trajectories (dashed lines) from learned dynamics.}
    \label{fig:set1}
\end{figure*}

Table \ref{tab:exp_set1} shows the ground truth equations and equations identified by using neural SINDy. We use $\PolySet_{2,3}$, $\PolySet_{3,3}$, and $\PolySet_{3,2}$, respectively, for the first three example problems, the fourth example problem, and the last example problem. During training, coefficients with small magnitude are pruned (i.e., setting them as 0) and are not presented in Table. We also report the learned coefficients without the pruning strategy in Appendix, which highlights that the pruning is required to zero out the coefficients of small magnitude. 

Figure \ref{fig:set1} depicts examples of reference trajectories and trajectories computed from identified dynamics, where the trajectories are chosen from the test set. As reported in the Appendix, time-instantaneous MSEs computed by nSINDy are several orders of magnitude smaller ($10^6\sim10^8$ smaller) than those computed by the black-box NODEs.

\section{Neural SINDy for structure preserving parameterization}

\subsection{GENERIC structure-preserving parameterization}
In the following, we present neural SINDy for a structure-preserving reversible and irreversible dynamics modeling approach. An alternative formalism for irreversible dynamics based on port-Hamiltonians is included in the appendix.

\paragraph{GENERIC formalism}
We begin by reviewing the general equation for the nonequilibrium reversible–irreversible coupling (GENERIC) formalism.

\begin{table*}[t]
\centering
\begin{tabular}{l|c|l|l}
    \hline
    Eq. name & Parameterization & Ground truth & Identified \\
    \hline
    \multirow{4}{*}{DNO} & \multirow{3}{*}{nSINDy} &  $\dot{q} = p$ & $\dot{q} =   0.999622 p$\\
    & & $\dot{p} = -3 \sin(q) - 0.04 p$ & $\dot{p} = -2.8645 \sin(q) -0.04001 p-0.1340 q +0.0190 q^3$\\ 
    & & $\dot{S} = 0.04p^2$ & $\dot{S} = 0.040081 p^2 $\\
    \cline{2-4}
    & nSINDy - GNN & $\TotalEnergy = 0.5 p^2 - 3 \cos (q) + S$ & $\TotalEnergyNN = 0.500144 p^2  -2.999785 \cos (q) + 0.998337 S$ \\
    \hline
\end{tabular}
\caption{[GENERIC structure preservation] The  equation names, the ground truth equations, and the identified equations by using nSINDy (non structure preserving parameterization) and nSINDy - GNN (GENERIC structure preserving parameterization).}
\label{tab:exp_set3}
\end{table*}

In GENERIC, the evolution of an observable $A(\State)$ is assumed to evolve under the gradient flow 
\begin{equation}
    \dv{A}{\TimeSymb} = \PoissonBracket{A}{\TotalEnergy} + \IrrBracket{A}{\Entropy}
\end{equation}
where $\TotalEnergy$ and $\Entropy$ denote generalized energy and entropy, and  $\PoissonBracket{\cdot}{\cdot}$ and $\IrrBracket{\cdot}{\cdot}$ denote a Poisson bracket and an irreversible metric bracket. The Poisson bracket is given in terms of a skew-symmetric Poisson matrix $\PoissonMatrix$ and the irreversible bracket is given in terms of a symmetric positive semi-definite friction matrix $\FrictionMatrix$,
\begin{equation}\label{eq:brackets}
    \PoissonBracket{A}{B} = \PoissonBracketBi{A}{B},\quad \mathrm{and} \quad \IrrBracket{A}{B} = \IrrBracketBi{A}{B}.
\end{equation}
Lastly, the GENERIC formalism requires two degeneracy conditions defined as  
\begin{equation}\label{eq:degeneracy_cond}
    \PoissonMatrix \pdv{\Entropy}{\State} = 0, \quad \mathrm{and} \quad \FrictionMatrix \pdv{\TotalEnergy}{\State}=0.
\end{equation}
With the state variables $\State = [q,p,\Entropy]\Transpose$, the GENERIC formalism defines the evolution of $\State$ as
\begin{equation}\label{eq:generic}
    \dv{\State}{\TimeSymb} = \PoissonMatrix \pdv{\TotalEnergy}{\State} +  \FrictionMatrix \pdv{\Entropy}{\State}.
\end{equation}

\paragraph{Parameterization for GENERIC}
Here, we review the parameterization technique, developed in \cite{oettinger2014irreversible} and further extended into deep learning settings in \cite{lee2021machine}. As for  the Hamiltonian Eq.~\eqref{eq:hamiltonian}, we parameterize the total energy such as
\begin{equation}
    \TotalEnergyNN(q,p,\Entropy) = (\DictionaryVec(q,p,\Entropy)\Transpose \CoeffsMat)\Transpose,
\end{equation}
with $\theta_1 = \CoeffsMat$. Then we parameterize the symmetric irreversible dynamics via the bracket 
\begin{equation}
    \IrrBracket{A}{B}_{\theta_2} = \zeta_{\alpha\beta,\mu\nu} \pdv{A}{x_\alpha} \pdv{\TotalEnergyNN}{x_\beta} \pdv{B}{x_\mu} \pdv{\TotalEnergyNN}{x_\nu},
\end{equation}
where
\begin{equation}
    \zeta_{\alpha\beta,\mu\nu} = \Lambda_{\alpha\beta}^m D_{mn} \Lambda_{\mu\nu}^n.
\end{equation}
The matrices, $\Lambda$ and $D$, are skew-symmetric and symmetric positive semi-definite matrices, respectively, such that
\begin{equation}
    \Lambda_{\alpha\beta}^m = -\Lambda_{\beta\alpha}^m, 
    \quad \mathrm{and} \quad
    D_{mn} = D_{nm},
\end{equation}
where the skew-symmetry and the symmetric positive semi-definiteness can be achieved by the parameterization tricks
\begin{equation}
    \Lambda = \frac{1}{2}(\tilde \Lambda - \tilde \Lambda\Transpose), \quad \mathrm{and} \quad \quad D = \tilde D \tilde D\Transpose,
\end{equation}
where $\tilde{\Lambda}$ and $\tilde{D}$ are matrices with learnable entries and, thus, $\theta_2 = [\tilde \Lambda, \tilde D]$.

With this parameterization, the irreversible part of the dynamics is given by 
\begin{equation}
    \IrrBracket{\State}{\Entropy}_{\theta_2} = \FrictionMatrix_{\theta_2}\pdv{\Entropy}{\State}= \zeta_{\alpha\beta,\mu\nu}  \pdv{\TotalEnergyNN}{x_\beta} \pdv{\Entropy}{x_\mu} \pdv{\TotalEnergyNN}{x_\nu}
\end{equation}
and, as a result, $\VelocityNN$ is now defined as 
\begin{equation}
    \VelocityNN = \PoissonBracket{\State}{\TotalEnergyNN} + \IrrBracket{\State}{\Entropy} = \PoissonMatrix\pdv{\TotalEnergyNN}{\State} + \FrictionMatrix_{\theta_2} \pdv{\Entropy}{\State},
\end{equation}
with $\NNParams = [\theta_1, \theta_2]$.

\begin{figure}[tb]
    \centering
     \includegraphics[scale=.55]{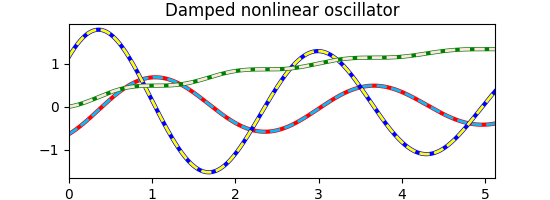}
    \caption{[GENERIC structure preservation] Examples of reference trajectories (solid lines) and computed trajectories (dashed lines) from learned dynamics.}
    \label{fig:set3}
\end{figure}

\begin{figure}[t]
    \centering
     \includegraphics[scale=.55]{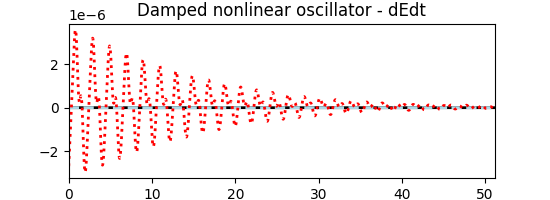}\\
     \includegraphics[scale=.55]{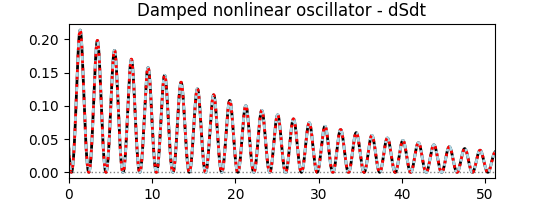}
     \includegraphics[scale=.55]{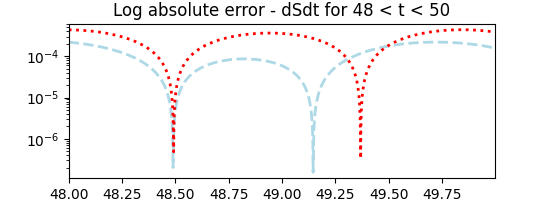}
    \caption{[GENERIC structure preservation] $\dv{\TotalEnergy}{\TimeSymb}$ and $\dv{\Entropy}{\TimeSymb}$of the reference system (black solid line), the system identified by nSINDy (red dotted line), and the system identified by nSINDy - GNN (lightblue dashed line).}
    \label{fig:set3_ES}
\end{figure}

\paragraph{Experiments}
An example problem (with their reference mathematical formulations) considered in this paper is:
\begin{itemize}
    \item Damped nonlinear oscillator from \cite{shang2020structure}: the ground truth equation can be written in the GENERIC formalism (Eq.~\eqref{eq:generic}) with the following components: 
    \begin{equation}
        \PoissonMatrix = \begin{bmatrix} 0 & 1 & 0\\ 
        -1 & 0 & 0\\
        0 & 0 & 0
        \end{bmatrix},\quad \FrictionMatrix = 0.04 \begin{bmatrix} 
        0 & 0 & 0 \\
        0 & 1 & -p\\
        0 & -p & p^2
        \end{bmatrix},
    \end{equation}
    \begin{equation}
        \text{and} \quad \TotalEnergy(q,p,S) = \frac{p^2}{2} - 3 \cos (q) + S.
    \end{equation}
    Alternatively, the equation for the dynamics can be written as 
    \begin{equation}
        \begin{split}
            \dot{q} &= p,\\
            \dot{p} &= -3 \sin(q) - 0.04 p \\
            \dot{S} &= 0.04 p^2.
        \end{split}
    \end{equation}
\end{itemize}

\begin{table*}[t]
\centering
\begin{tabular}{l|c|l|l}
    \hline
    Eq. name & Parameterization & Ground truth & Identified \\
    \hline
    \multirow{3}{*}{Ideal mass-spring}  & \multirow{2}{*}{nSINDy} & $\dot{q} = p$ & $\dot{q} =   0.999989 p$\\ 
     & & $\dot{p} = -q$ & $\dot{p} = -1.000030 q$\\
     \cline{2-4}
     & nSINDy - HNN & $\Hamiltonian(q,p) = 0.5 q^2 + 0.5p^2$& $\Hamiltonian(q,p) = 0.499991 q^2 + 0.500033p^2$\\
    \hline
    \multirow{3}{*}{Ideal pendulum} & \multirow{2}{*}{nSINDy}  & $\dot{q} = p$ & $\dot{q} =   1.000018 p + 0.034095 q -0.004796 q^3$\\
    & & $\dot{p} = - 6 \sin(q) $ & $\dot{p} = -5.890695 \sin (q)  -0.107052q$ \\ 
    \cline{2-4}
     & nSINDy - HNN & $\Hamiltonian(q,p) = 6- 6 \cos( q) + 0.5 p^2$ & $\Hamiltonian(q,p) = 6-5.999838\cos( q) + 0.500014 p^2$\\
    \hline
\end{tabular}
\caption{[Hamiltonian structure preservation] The  equation names, the ground truth equations, and the identified equations by using nSINDy (non structure preserving parameterization) and nSINDy - HNN (Hamiltonian structure preserving parameterization).}
\label{tab:exp_set2}
\end{table*}

In the experiment, we assume that we have knowledge on $\PoissonMatrix$ and that the measurements on $\Entropy$ are available. That is, the nSINDy  method with the GENERIC structure preservation seeks the unknown $\FrictionMatrix$ and $\TotalEnergy$. For generating data, we base our implementation on the code from \cite{shang2020structure}. We generate 800 training trajectories, 160 validation trajectories, and 160 test trajectories with $\Delta t = 0.001$ and the simulation time $[0,5.12]$. We use a dictionary consisting of polynomials and trigonometric functions:
\begin{equation}
    \PolySet = \{\PolySet_{2,3}, \cos(q),\sin(q),\cos(p),\sin(p)\}.
\end{equation}
We consider the same experimental settings that are used in the above experiments: Adamax optimizer with the initial learning rate 0.01, exponential learning rate decay with a factor 0.9987, L1-penalty weight $10^{-4}$, pruning threshold $10^{-6}$, dopri5 for the ODE integrator with $10^{-7}$ and $10^{-9}$ for relative and absolute tolerances, and, finally, $n_{\max}=300$.

Table \ref{tab:exp_set3} reports the coefficients of the identified systems when nSINDy and nSINDy - GNN are used. With $n_{\max}=300$, nSINDy fails to correctly identify the system, whereas nSINDy - GNN identifies the exact terms correctly and computes the coefficients accurately. Figure \ref{fig:set3} depicts the ground truth trajectory and the trajectory computed from the learned dynamics function (nSINDy - GNN). Figure \ref{fig:set3_ES} shows the advantages of using GENERIC structure preservation via plots of  $\dv{\TotalEnergy}{t}$
and $\dv{\Entropy}{t}$. The difference between structure-preserving and non-structure-preserving parameterization is shown dramatically in the plot of $\dv{\TotalEnergy}{t}$; the structure-preserving parameterization produces $\dv{\TotalEnergy}{t}=0$, whereas the non-structure-preserving parameterization produces fluctuating values of $\dv{\TotalEnergy}{t}$.

\begin{figure}[tb]
    \centering
     \includegraphics[scale=.55]{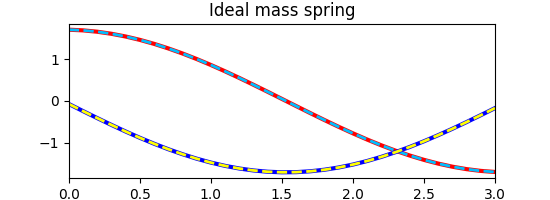}\\
    \includegraphics[scale=.55]{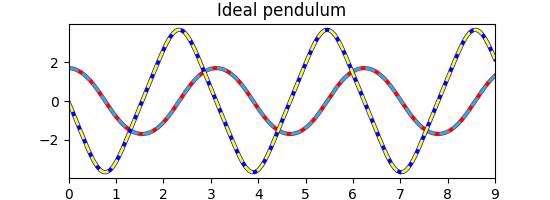}
    \caption{[Hamiltonian structure preservation]Examples of reference trajectories (solid lines) and computed trajectories (dashed lines) from learned dynamics.}
    \label{fig:set2}
\end{figure}

\begin{figure}[tb]
    \centering
    \includegraphics[scale=.55]{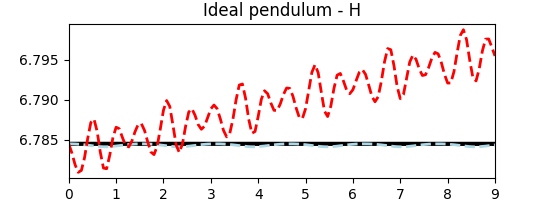}\\
    \includegraphics[scale=.55]{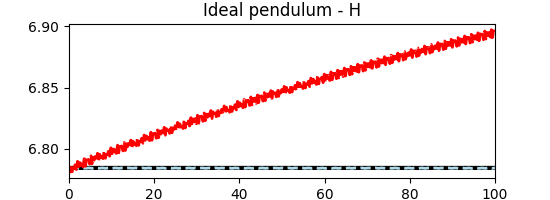}
    \caption{[Hamiltonian structure preservation] The Hamiltonian function measured by the trajectories computed with nSINDy (red dashed line) and nSINDy-HNN (blue dashed line). Figure below depicts the Hamiltonian functions measured by the trajectories computed for extended numerical simulation time, i.e., 100 seconds (the training simulation time is 9 seconds).}
    \label{fig:set2_H}
\end{figure}

\subsection{Hamiltonian structure-preserving parameterization}
In the following, we consider the Hamiltonian structure-preserving parameterization technique proposed in Hamiltonian neural networks (HNNs) \cite{Greydanus2019hnn}: parameterizing the Hamiltonian function $\Hamiltonian(q,p)$ as $\Hamiltonian_{\NNParams}(q,p)$ such that 
\begin{equation}\label{eq:hamiltonian}
    \Hamiltonian_{\NNParams} = (\DictionaryVec(q,p)\Transpose \CoeffsMat)\Transpose.
\end{equation}
With the above parameterization and $\State = [q,p]\Transpose$, the dynamics can be modeled as 
\begin{equation}
    \begin{bmatrix}
    \dv{q}{\TimeSymb}\\[8pt]
    \dv{p}{\TimeSymb}
    \end{bmatrix} 
    = \begin{bmatrix*}[r]
    -\pdv{\Hamiltonian_{\NNParams}}{p}\\[8pt]
    \pdv{\Hamiltonian_{\NNParams}}{q}
    \end{bmatrix*},
\end{equation}
i.e., $\VelocityNN$ is now defined as $\VelocityNN = [-\pdv{\Hamiltonian_{\NNParams}}{p}, \pdv{\Hamiltonian_{\NNParams}}{q}]\Transpose$.
 Then the loss objective can be written as 
\begin{equation}
    \left \| \pdv{\Hamiltonian_{\NNParams}}{p} - \dv{q}{\TimeSymb} \right\|_1 + \left \| \pdv{\Hamiltonian_{\NNParams}}{q} + \dv{p}{\TimeSymb} \right\|_1. 
\end{equation}

As noted in \cite{lee2021machine}, 
with canonical coordinates $\State = [q,p]\Transpose$, and canonical Poisson matrix $\PoissonMatrix= \begin{bmatrix}0 & 1 \\ -1 & 0\end{bmatrix}$, and $\FrictionMatrix = \bm{0}$, the GENERIC formalism in Eq.~\eqref{eq:generic} recovers Hamiltonian dynamics.

\paragraph{Experiments}
We examine the performance of the proposed method with a list of example dynamic systems:
\begin{itemize}
    \item Ideal mass-spring,
    \item Ideal pendulum.
\end{itemize}
For generating data, we base our implementation on the code from \cite{Greydanus2019hnn}. We generate 800 training trajectories, 160 validation trajectories, and 160 test trajectories. For the mass spring and the pendulum problem, we use $\Delta t = 0.1$ and $\Delta t = 1/15$, respectively, and the simulation time is set as $[0,3]$ and $[0,9]$, respectively.

Table \ref{tab:exp_set2} shows the ground truth equations and equations identified by using neural SINDy. For the mass spring problem, we use $\PolySet_{2,3}$ and for the pendulum problem we use a dictionary consisting of polynomials and trigonometric functions:
\begin{equation}
    \PolySet = \{\PolySet_{2,3}, \cos(q),\sin(q),\cos(p),\sin(p)\}.
\end{equation} 
Again, we use the same experimental settings described above for the GENERIC parameterization. In Table \ref{tab:exp_set2}, we presented results obtained by two different  parameterization techniques: 1) the ``plain'' dictionary approach (Eq.~\eqref{eq:dict}) denoted by (nSINDy) and 2) the Hamiltonian approach (Eq~\eqref{eq:hamiltonian}) denoted by (nSINDy - HNN).

Figure \ref{fig:set2} depicts examples of reference trajectories and trajectories computed from identified dynamics, where the trajectories are chosen from the test set. Figure \ref{fig:set2_H} depicts how the energy is being conserved in the dynamics learned with nSINDy-HNN, whereas the plain apporach (nSINDy) fails to conserve the energy.

\section{Discussion}
\paragraph{Limitation}
The proposed method shares the same limitations of the original SINDy method: successful identification requires inclusion of the correct dictionaries in the library. Potential alternatives are well-studied in the literature and include either adding an extra  ``black-box'' neural network to compensate missing dictionaries or designing a neural network that can learn dictionaries from data (such as in \cite{sahoo2018learning}).

The gradient-based parameter update and the magnitude-based pruning could potentially zero out unwanted coefficients in a special case: when the signs of the coefficients need to be changed in the later stage of the training. If the magnitude of the loss term becomes very small (and the gradient as well), the updated coefficients may satisfy the pruning condition shown in \eqref{eq:prune}.

Lastly, when adaptive step-size ODE solvers are used (e.g., dopri5), numerical underflow may occur. This can be mitigated by trying different initialization for the coefficients, or smaller batch length $\LBatch$. However, further study regarding robust initialization is required.

\section{Related work}
\paragraph{System identification}
In \cite{schmidt2009distilling}, the proposed method uses a genetic algorithms to identify governing physical laws (Lagrangian or Hamiltonian) from measurements of real experiments. A seminal work on sparse regression methods for system identification has been proposed in \cite{brunton2016discovering}. Then the sparse regression methods have been extended to various settings, e.g., for model predictive control \cite{kaiser2018sparse}, and for identifying dynamics in latent space using autoencoders \cite{hinton2006reducing} and then learning parsimonious representations for the latent dynamics \cite{champion2019data}. Also, the sparse regression methods have been applied for identifying partial differential equations (PDEs) \cite{rudy2017data,rudy2019data}.  For identifying PDEs,  deep-learning-based approaches such as physics-informed neural networks \cite{raissi2019physics} and PDE-net \cite{long2018pde} have been studied recently.

\paragraph{Structure preserving neural network} Designing neural network architectures that exactly enforces important physical properties has been an important topic and studied extensively. Parameterization techniques that preserve physical structure include Hamiltonian neural networks \cite{Greydanus2019hnn,toth2019hamiltonian}, Lagrangian neural networks \cite{cranmer2020lagrangian,lutter2018deep}, port-Hamiltonian neural networks \cite{desai2021port}, and GENERIC neural networks \cite{hernandez2021structure,lee2021machine}. Neural network architectures that mimic the action of symplectic integrators have been proposed in \cite{chen2019symplectic,jin2020sympnets,tong2021symplectic}, and a training algorithm exploiting physical invariance for learning dynamical systems, e.g., time-reversal symmetric, has been studied in \cite{huh2020time}.

\section{Conclusion}
We have proposed a simple and effective deep-learning-based training algorithm for a dictionary-based parameterization of nonlinear dynamics. The proposed algorithm is based on the training procedure introduced in neural ordinary differential equations (NODE) \cite{chen2018neural} and employs L1-weight decay of  model parameters and magnitude-based pruning strategy, inspired by sparse identification of nonlinear dynamics (SINDy) \cite{brunton2016discovering}. We have further extended the dictionary-based parameterization approach to structure-preserving parameterization techniques, such as Hamiltonian neural networks, GENERIC neural networks, and port-Hamiltonian networks. For a suite of benchmark problems, we have demonstrated that the proposed training algorithm is very effective in identifying the underlying dynamics from data with expected gains from imposing structure-preservation.

\section{Acknowledgments}
\label{sec:acknowledge}
N.~Trask and P.~Stinis acknowledge funding under the Collaboratory on Mathematics and Physics-Informed Learning Machines for Multiscale and Multiphysics Problems (PhILMs) project funded by DOE Office of Science (Grant number DE-SC001924). N.~Trask and K.~Lee acknowledge funding from the DOE Early Career program. Sandia National Laboratories is a multi-mission laboratory managed and operated by National Technology and Engineering Solutions of Sandia, LLC., a wholly owned subsidiary of Honeywell International, Inc., for the U.S. Department of Energy’s National Nuclear Security Administration under contract DE-NA0003525. This paper describes objective technical results and analysis.  Any subjective views or opinions that might be expressed in the paper do not necessarily represent the views of the U.S. Department of Energy or the United States Government.

\bibliography{aaai22.bib}

\begin{thebibliography}{50}
\providecommand{\natexlab}[1]{#1}

\bibitem[{Alexopoulos, Nikolakis, and
  Chryssolouris(2020)}]{alexopoulos2020digital}
Alexopoulos, K.; Nikolakis, N.; and Chryssolouris, G. 2020.
\newblock Digital twin-driven supervised machine learning for the development
  of artificial intelligence applications in manufacturing.
\newblock \emph{International Journal of Computer Integrated Manufacturing},
  33(5): 429--439.

\bibitem[{Baker et~al.(2019)Baker, Alexander, Bremer, Hagberg, Kevrekidis,
  Najm, Parashar, Patra, Sethian, Wild et~al.}]{baker2019workshop}
Baker, N.; Alexander, F.; Bremer, T.; Hagberg, A.; Kevrekidis, Y.; Najm, H.;
  Parashar, M.; Patra, A.; Sethian, J.; Wild, S.; et~al. 2019.
\newblock Workshop report on basic research needs for scientific machine
  learning: Core technologies for artificial intelligence.
\newblock Technical report, USDOE Office of Science (SC), Washington, DC
  (United States).

\bibitem[{Bertalan et~al.(2019)Bertalan, Dietrich, Mezi{\'c}, and
  Kevrekidis}]{bertalan2019learning}
Bertalan, T.; Dietrich, F.; Mezi{\'c}, I.; and Kevrekidis, I.~G. 2019.
\newblock On learning Hamiltonian systems from data.
\newblock \emph{Chaos: An Interdisciplinary Journal of Nonlinear Science},
  29(12): 121107.

\bibitem[{Brunton, Proctor, and Kutz(2016)}]{brunton2016discovering}
Brunton, S.~L.; Proctor, J.~L.; and Kutz, J.~N. 2016.
\newblock Discovering governing equations from data by sparse identification of
  nonlinear dynamical systems.
\newblock \emph{Proceedings of the National Academy of Sciences}, 113(15):
  3932--3937.

\bibitem[{Champion et~al.(2019)Champion, Lusch, Kutz, and
  Brunton}]{champion2019data}
Champion, K.; Lusch, B.; Kutz, J.~N.; and Brunton, S.~L. 2019.
\newblock Data-driven discovery of coordinates and governing equations.
\newblock \emph{Proceedings of the National Academy of Sciences}, 116(45):
  22445--22451.

\bibitem[{Chen and Tao(2021)}]{chen2021data}
Chen, R.; and Tao, M. 2021.
\newblock Data-driven Prediction of General Hamiltonian Dynamics via Learning
  Exactly-Symplectic Maps.
\newblock \emph{arXiv preprint arXiv:2103.05632}.

\bibitem[{Chen et~al.(2018)Chen, Rubanova, Bettencourt, and
  Duvenaud}]{chen2018neural}
Chen, R.~T.; Rubanova, Y.; Bettencourt, J.; and Duvenaud, D. 2018.
\newblock Neural ordinary differential equations.
\newblock In \emph{Proceedings of the 32nd International Conference on Neural
  Information Processing Systems}, 6572--6583.

\bibitem[{Chen et~al.(2021)Chen, Wang, Hesthaven, and Zhang}]{chen2021physics}
Chen, W.; Wang, Q.; Hesthaven, J.~S.; and Zhang, C. 2021.
\newblock Physics-informed machine learning for reduced-order modeling of
  nonlinear problems.
\newblock \emph{Journal of Computational Physics}, 110666.

\bibitem[{Chen et~al.(2019)Chen, Zhang, Arjovsky, and
  Bottou}]{chen2019symplectic}
Chen, Z.; Zhang, J.; Arjovsky, M.; and Bottou, L. 2019.
\newblock Symplectic Recurrent Neural Networks.
\newblock In \emph{International Conference on Learning Representations}.

\bibitem[{Cranmer et~al.(2020)Cranmer, Greydanus, Hoyer, Battaglia, Spergel,
  and Ho}]{cranmer2020lagrangian}
Cranmer, M.; Greydanus, S.; Hoyer, S.; Battaglia, P.; Spergel, D.; and Ho, S.
  2020.
\newblock Lagrangian Neural Networks.
\newblock In \emph{ICLR 2020 Workshop on Integration of Deep Neural Models and
  Differential Equations}.

\bibitem[{Desai et~al.(2021)Desai, Mattheakis, Sondak, Protopapas, and
  Roberts}]{desai2021port}
Desai, S.; Mattheakis, M.; Sondak, D.; Protopapas, P.; and Roberts, S. 2021.
\newblock Port-Hamiltonian Neural Networks for Learning Explicit Time-Dependent
  Dynamical Systems.
\newblock \emph{arXiv preprint arXiv:2107.08024}.

\bibitem[{Dormand and Prince(1980)}]{dormand1980family}
Dormand, J.~R.; and Prince, P.~J. 1980.
\newblock A family of embedded {R}unge-{K}utta formulae.
\newblock \emph{Journal of computational and applied mathematics}, 6(1):
  19--26.

\bibitem[{Ghnatios et~al.(2019)Ghnatios, Alfaro, Gonz{\'a}lez, Chinesta, and
  Cueto}]{ghnatios2019data}
Ghnatios, C.; Alfaro, I.; Gonz{\'a}lez, D.; Chinesta, F.; and Cueto, E. 2019.
\newblock Data-driven generic modeling of poroviscoelastic materials.
\newblock \emph{Entropy}, 21(12): 1165.

\bibitem[{Greydanus, Dzamba, and Yosinski(2019)}]{Greydanus2019hnn}
Greydanus, S.; Dzamba, M.; and Yosinski, J. 2019.
\newblock Hamiltonian Neural Networks.
\newblock In Wallach, H.~M.; Larochelle, H.; Beygelzimer, A.;
  d'Alch{\'{e}}{-}Buc, F.; Fox, E.~B.; and Garnett, R., eds., \emph{Advances in
  Neural Information Processing Systems 32}, 15353--15363.

\bibitem[{Hairer et~al.(2006)Hairer, Hochbruck, Iserles, and
  Lubich}]{hairer2006geometric}
Hairer, E.; Hochbruck, M.; Iserles, A.; and Lubich, C. 2006.
\newblock Geometric numerical integration.
\newblock \emph{Oberwolfach Reports}, 3(1): 805--882.

\bibitem[{Hern{\'a}ndez et~al.(2021)Hern{\'a}ndez, Bad{\'\i}as, Gonz{\'a}lez,
  Chinesta, and Cueto}]{hernandez2021structure}
Hern{\'a}ndez, Q.; Bad{\'\i}as, A.; Gonz{\'a}lez, D.; Chinesta, F.; and Cueto,
  E. 2021.
\newblock Structure-preserving neural networks.
\newblock \emph{Journal of Computational Physics}, 426: 109950.

\bibitem[{Hinton and Salakhutdinov(2006)}]{hinton2006reducing}
Hinton, G.~E.; and Salakhutdinov, R.~R. 2006.
\newblock Reducing the dimensionality of data with neural networks.
\newblock \emph{science}, 313(5786): 504--507.

\bibitem[{Huh et~al.(2020)Huh, Yang, Hwang, and Shin}]{huh2020time}
Huh, I.; Yang, E.; Hwang, S.~J.; and Shin, J. 2020.
\newblock Time-Reversal Symmetric ODE Network.
\newblock \emph{Advances in Neural Information Processing Systems}, 33.

\bibitem[{Jin et~al.(2020)Jin, Zhang, Zhu, Tang, and
  Karniadakis}]{jin2020sympnets}
Jin, P.; Zhang, Z.; Zhu, A.; Tang, Y.; and Karniadakis, G.~E. 2020.
\newblock SympNets: Intrinsic structure-preserving symplectic networks for
  identifying Hamiltonian systems.
\newblock \emph{Neural Networks}, 132: 166--179.

\bibitem[{Kaiser, Kutz, and Brunton(2018)}]{kaiser2018sparse}
Kaiser, E.; Kutz, J.~N.; and Brunton, S.~L. 2018.
\newblock Sparse identification of nonlinear dynamics for model predictive
  control in the low-data limit.
\newblock \emph{Proceedings of the Royal Society A}, 474(2219): 20180335.

\bibitem[{Karapiperis et~al.(2021)Karapiperis, Stainier, Ortiz, and
  Andrade}]{karapiperis2021data}
Karapiperis, K.; Stainier, L.; Ortiz, M.; and Andrade, J. 2021.
\newblock Data-driven multiscale modeling in mechanics.
\newblock \emph{Journal of the Mechanics and Physics of Solids}, 147: 104239.

\bibitem[{Karniadakis et~al.(2021)Karniadakis, Kevrekidis, Lu, Perdikaris,
  Wang, and Yang}]{karniadakis2021physics}
Karniadakis, G.~E.; Kevrekidis, I.~G.; Lu, L.; Perdikaris, P.; Wang, S.; and
  Yang, L. 2021.
\newblock Physics-informed machine learning.
\newblock \emph{Nature Reviews Physics}, 3(6): 422--440.

\bibitem[{Kingma and Ba(2015)}]{kingma2015Adam}
Kingma, D.~P.; and Ba, J. 2015.
\newblock Adam: {A} Method for Stochastic Optimization.
\newblock In \emph{3rd International Conference on Learning Representations,
  {ICLR}}.

\bibitem[{Lee and Carlberg(2021)}]{lee2021deep}
Lee, K.; and Carlberg, K.~T. 2021.
\newblock Deep Conservation: A Latent-Dynamics Model for Exact Satisfaction of
  Physical Conservation Laws.
\newblock In \emph{Proceedings of the AAAI Conference on Artificial
  Intelligence}, volume~35, 277--285.

\bibitem[{Lee, Trask, and Stinis(2021)}]{lee2021machine}
Lee, K.; Trask, N.~A.; and Stinis, P. 2021.
\newblock Machine learning structure preserving brackets for forecasting
  irreversible processes.
\newblock \emph{arXiv preprint arXiv:2106.12619}.

\bibitem[{Liu, Kutz, and Brunton(2020)}]{liu2020hierarchical}
Liu, Y.; Kutz, J.~N.; and Brunton, S.~L. 2020.
\newblock Hierarchical deep learning of multiscale differential equation
  time-steppers.
\newblock \emph{arXiv preprint arXiv:2008.09768}.

\bibitem[{Long et~al.(2018)Long, Lu, Ma, and Dong}]{long2018pde}
Long, Z.; Lu, Y.; Ma, X.; and Dong, B. 2018.
\newblock Pde-net: Learning pdes from data.
\newblock In \emph{International Conference on Machine Learning}, 3208--3216.
  PMLR.

\bibitem[{Lutter, Ritter, and Peters(2018)}]{lutter2018deep}
Lutter, M.; Ritter, C.; and Peters, J. 2018.
\newblock Deep Lagrangian Networks: Using Physics as Model Prior for Deep
  Learning.
\newblock In \emph{International Conference on Learning Representations}.

\bibitem[{Marsden and Ratiu(1995)}]{marsden1995introduction}
Marsden, J.~E.; and Ratiu, T.~S. 1995.
\newblock Introduction to mechanics and symmetry.
\newblock \emph{Physics Today}, 48(12): 65.

\bibitem[{Masi et~al.(2021)Masi, Stefanou, Vannucci, and
  Maffi-Berthier}]{masi2021thermodynamics}
Masi, F.; Stefanou, I.; Vannucci, P.; and Maffi-Berthier, V. 2021.
\newblock Thermodynamics-based Artificial Neural Networks for constitutive
  modeling.
\newblock \emph{Journal of the Mechanics and Physics of Solids}, 147: 104277.

\bibitem[{Oettinger(2014)}]{oettinger2014irreversible}
Oettinger, H.~C. 2014.
\newblock Irreversible dynamics, {O}nsager--{C}asimir symmetry, and an
  application to turbulence.
\newblock \emph{Physical Review E}, 90(4): 042121.

\bibitem[{Paszke et~al.(2019)Paszke, Gross, Massa, Lerer, Bradbury, Chanan,
  Killeen, Lin, Gimelshein, Antiga, Desmaison, K{\"{o}}pf, Yang, DeVito,
  Raison, Tejani, Chilamkurthy, Steiner, Fang, Bai, and
  Chintala}]{paszke2019pytorch}
Paszke, A.; Gross, S.; Massa, F.; Lerer, A.; Bradbury, J.; Chanan, G.; Killeen,
  T.; Lin, Z.; Gimelshein, N.; Antiga, L.; Desmaison, A.; K{\"{o}}pf, A.; Yang,
  E.; DeVito, Z.; Raison, M.; Tejani, A.; Chilamkurthy, S.; Steiner, B.; Fang,
  L.; Bai, J.; and Chintala, S. 2019.
\newblock Py{T}orch: An Imperative Style, High-Performance Deep Learning
  Library.
\newblock In \emph{Advances in Neural Information Processing Systems 32},
  8024--8035.

\bibitem[{Patel et~al.(2020)Patel, Manickam, Trask, Wood, Lee, Tomas, and
  Cyr}]{patel2020thermodynamically}
Patel, R.~G.; Manickam, I.; Trask, N.~A.; Wood, M.~A.; Lee, M.; Tomas, I.; and
  Cyr, E.~C. 2020.
\newblock Thermodynamically consistent physics-informed neural networks for
  hyperbolic systems.
\newblock \emph{arXiv preprint arXiv:2012.05343}.

\bibitem[{Raissi, Perdikaris, and Karniadakis(2019)}]{raissi2019physics}
Raissi, M.; Perdikaris, P.; and Karniadakis, G.~E. 2019.
\newblock Physics-informed neural networks: A deep learning framework for
  solving forward and inverse problems involving nonlinear partial differential
  equations.
\newblock \emph{Journal of Computational Physics}, 378: 686--707.

\bibitem[{Rohrhofer, Posch, and Geiger(2021)}]{rohrhofer2021pareto}
Rohrhofer, F.~M.; Posch, S.; and Geiger, B.~C. 2021.
\newblock On the Pareto Front of Physics-Informed Neural Networks.
\newblock \emph{arXiv preprint arXiv:2105.00862}.

\bibitem[{Rudy et~al.(2019)Rudy, Alla, Brunton, and Kutz}]{rudy2019data}
Rudy, S.; Alla, A.; Brunton, S.~L.; and Kutz, J.~N. 2019.
\newblock Data-driven identification of parametric partial differential
  equations.
\newblock \emph{SIAM Journal on Applied Dynamical Systems}, 18(2): 643--660.

\bibitem[{Rudy et~al.(2017)Rudy, Brunton, Proctor, and Kutz}]{rudy2017data}
Rudy, S.~H.; Brunton, S.~L.; Proctor, J.~L.; and Kutz, J.~N. 2017.
\newblock Data-driven discovery of partial differential equations.
\newblock \emph{Science Advances}, 3(4): e1602614.

\bibitem[{Sahoo, Lampert, and Martius(2018)}]{sahoo2018learning}
Sahoo, S.; Lampert, C.; and Martius, G. 2018.
\newblock Learning equations for extrapolation and control.
\newblock In \emph{International Conference on Machine Learning}, 4442--4450.
  PMLR.

\bibitem[{Schmidt and Lipson(2009)}]{schmidt2009distilling}
Schmidt, M.; and Lipson, H. 2009.
\newblock Distilling free-form natural laws from experimental data.
\newblock \emph{science}, 324(5923): 81--85.

\bibitem[{Shang and {\"O}ttinger(2020)}]{shang2020structure}
Shang, X.; and {\"O}ttinger, H.~C. 2020.
\newblock Structure-preserving integrators for dissipative systems based on
  reversible--irreversible splitting.
\newblock \emph{Proceedings of the Royal Society A}, 476(2234): 20190446.

\bibitem[{Tibshirani(1996)}]{tibshirani1996regression}
Tibshirani, R. 1996.
\newblock Regression shrinkage and selection via the {L}asso.
\newblock \emph{Journal of the Royal Statistical Society: Series B
  (Methodological)}, 58(1): 267--288.

\bibitem[{Tong et~al.(2021)Tong, Xiong, He, Pan, and Zhu}]{tong2021symplectic}
Tong, Y.; Xiong, S.; He, X.; Pan, G.; and Zhu, B. 2021.
\newblock Symplectic neural networks in Taylor series form for Hamiltonian
  systems.
\newblock \emph{Journal of Computational Physics}, 110325.

\bibitem[{Toth et~al.(2019)Toth, Rezende, Jaegle, Racani{\`e}re, Botev, and
  Higgins}]{toth2019hamiltonian}
Toth, P.; Rezende, D.~J.; Jaegle, A.; Racani{\`e}re, S.; Botev, A.; and
  Higgins, I. 2019.
\newblock Hamiltonian Generative Networks.
\newblock In \emph{International Conference on Learning Representations}.

\bibitem[{Trask, Huang, and Hu(2020)}]{trask2020enforcing}
Trask, N.; Huang, A.; and Hu, X. 2020.
\newblock Enforcing exact physics in scientific machine learning: a data-driven
  exterior calculus on graphs.
\newblock \emph{arXiv preprint arXiv:2012.11799}.

\bibitem[{Wan et~al.(2018)Wan, Vlachas, Koumoutsakos, and Sapsis}]{wan2018data}
Wan, Z.~Y.; Vlachas, P.; Koumoutsakos, P.; and Sapsis, T. 2018.
\newblock Data-assisted reduced-order modeling of extreme events in complex
  dynamical systems.
\newblock \emph{PloS one}, 13(5): e0197704.

\bibitem[{Wang, Teng, and Perdikaris(2020)}]{wang2020understanding}
Wang, S.; Teng, Y.; and Perdikaris, P. 2020.
\newblock Understanding and mitigating gradient pathologies in physics-informed
  neural networks.
\newblock \emph{arXiv preprint arXiv:2001.04536}.

\bibitem[{Wu, Xiao, and Paterson(2018)}]{wu2018physics}
Wu, J.-L.; Xiao, H.; and Paterson, E. 2018.
\newblock Physics-informed machine learning approach for augmenting turbulence
  models: A comprehensive framework.
\newblock \emph{Physical Review Fluids}, 3(7): 074602.

\bibitem[{Yu et~al.(2020)Yu, Tian, Li et~al.}]{yu2020onsagernet}
Yu, H.; Tian, X.; Li, Q.; et~al. 2020.
\newblock OnsagerNet: Learning Stable and Interpretable Dynamics using a
  Generalized Onsager Principle.
\newblock \emph{arXiv preprint arXiv:2009.02327}.

\bibitem[{Zhong, Dey, and Chakraborty(2020)}]{zhong2020dissipative}
Zhong, Y.~D.; Dey, B.; and Chakraborty, A. 2020.
\newblock Dissipative symoden: Encoding Hamiltonian dynamics with dissipation
  and control into deep learning.
\newblock \emph{arXiv preprint arXiv:2002.08860}.

\bibitem[{Zhong, Dey, and Chakraborty(2021)}]{zhong2021benchmarking}
Zhong, Y.~D.; Dey, B.; and Chakraborty, A. 2021.
\newblock Benchmarking Energy-Conserving Neural Networks for Learning Dynamics
  from Data.
\newblock In \emph{Learning for Dynamics and Control}, 1218--1229. PMLR.

\end{thebibliography}
\clearpage
\onecolumn
\appendix
\section{Training without pruning}
Tables \ref{tab:exp_set_appendix} and \ref{tab:exp_set_appendix2} report the coefficients learned by training without the pruning-strategy. Although the algorithm successfully finds the significant coefficients, it fails to zero out the coefficients of the small contributions. 
\begin{table}[htb]
\centering
\resizebox{.7\columnwidth}{!}{
\begin{tabular}{l|r|r|r|r|r|r|}
    \hline
    & \multicolumn{2}{c|}{Hyperbolic} & \multicolumn{2}{c|}{Cubic oscillator} & \multicolumn{2}{c|}{Van der Pol} \\
    \cline{2-7}
    & \multicolumn{2}{l|}{$\dot{x} = \Red{-0.05}x$} & \multicolumn{2}{l|}{$\dot{x} = \Red{-0.1}x^3 + \Blue{2}y^3$} & \multicolumn{2}{l|}{$\dot{x} = \Red{1}y$}\\
    & \multicolumn{2}{l|}{$\dot{y} = \Red{1}x^2\Blue{-1}y$} & \multicolumn{2}{l|}{$\dot{y} = \Red{-2}x^3\Blue{-0.1}y^3$} & \multicolumn{2}{l|}{$\dot{y} = \Red{-1}x+\Blue{2}y\Green{-2}x^2y$} \\
    \hline
    & \multicolumn{1}{c|}{$\dot{x}$} & \multicolumn{1}{c|}{$\dot{y}$} & \multicolumn{1}{c|}{$\dot{x}$} & \multicolumn{1}{c|}{$\dot{y}$} & \multicolumn{1}{c|}{$\dot{x}$} & \multicolumn{1}{c|}{$\dot{y}$}\\
    \hline
    $1$     & -4.0265e-04   & -1.5801e-04 & -1.1837e-04 & -1.9252e-04 & 7.6617e-04 & -4.6053e-04\\
    $x$     & \Red{-4.8321e-02}   & 6.7640e-04 & -3.8775e-04 & 2.1818e-05 & -4.3242e-04 & \Red{-9.9930e-01}\\
    $y$     & 2.2759e-05    & \Blue{-9.9818e-01} & 3.4231e-04 & 5.1497e-04 & \Red{1.0002e+00} & \Blue{1.9997e+00} \\
    $x^2$   & 1.1598e-04    & \Red{1.0021e+00} & 8.5114e-04 & -1.0675e-04 & -4.3595e-06 & -2.6983e-04\\
    $xy$    & -9.0811e-04   & 1.3638e-03 & 3.9183e-04 & -1.0255e-04 & 1.3413e-03 & 1.3681e-04\\
    $y^2$   & -9.9513e-04   & 1.2584e-03 & 4.7589e-04 & -3.6600e-04 & 8.9859e-04 & -5.9787e-04\\
    $x^3$   & 2.3237e-03    & -1.3569e-03 & \Red{-9.9985e-02} & \Red{-2.0003e+00} & 6.0941e-05 & -5.1159e-05\\
    $x^2y$  & -7.3889e-04   & 1.1560e-03 & -8.8203e-04 & 8.3778e-04 & 1.6378e-04 & \Green{-2.0006e+00}\\
    $xy^2$  & -3.7055e-04   & 3.2972e-04 & -3.6603e-04 & -6.3260e-04 & 3.8560e-04 & 3.5658e-04\\
    $y^3$   & -2.8758e-03   & 3.6145e-04 & \Blue{2.0003e+00} & \Blue{-1.0056e-01} & 1.3837e-04 & 6.8370e-05\\
    \hline
\end{tabular}}
\caption{[Hyperbolic, cubic oscillator, and Van der Pol problems] The coefficients learned by nSINDy via training without the magnitude-based pruning strategy.}
\label{tab:exp_set_appendix}
\end{table}

\begin{table}[htb]
\centering
\resizebox{.7\columnwidth}{!}{
\begin{tabular}{l|r|r|r|l|r|r|r|}
    \hline
    & \multicolumn{3}{c|}{Hopf bifurcation} & & \multicolumn{3}{c|}{Lorenz} \\
    \cline{2-8}
    & \multicolumn{3}{l|}{$\dot{x} = \Red{1} x\mu + \Blue{1}y\Green{-1}x^3\Yellow{-1}xy^2$} & & \multicolumn{3}{l|}{$\dot{x} = \Red{-10}x + \Blue{10}y$}\\
    & \multicolumn{3}{l|}{$\dot{y} = \Red{1} y\mu\Blue{-1}x\Green{-1}x^2y\Yellow{-1}y^3$} & & \multicolumn{3}{l|}{$\dot{y} = \Red{28}x\Blue{-1}xz\Green{-1}y$} \\
    & \multicolumn{3}{l|}{$\dot{\mu} = \Red{0}$} & & \multicolumn{3}{l|}{$\dot{z} = \Red{1}xy \Blue{-8/3}z$} \\
    \hline
    & \multicolumn{1}{c|}{$\dot{x}$} & \multicolumn{1}{c|}{$\dot{y}$} & \multicolumn{1}{c|}{$\dot{\mu}$} & & \multicolumn{1}{c|}{$\dot{x}$} & \multicolumn{1}{c|}{$\dot{y}$} & 
    \multicolumn{1}{c|}{$\dot{z}$}\\
    \hline
    $1$     & -7.7859e-05 & -5.4043e-05 & \Red{7.5873e-05} & $1$ & 5.1030e-07 & -5.2767e-07 & 8.4725e-07\\
    $x$     & -9.0900e-05 & \Blue{-1.0005e+00} & -2.7873e-04 & $x$ & \Red{5.6920e-01} & \Red{1.1720e+01} & -3.6492e-07\\
    $y$     & \Blue{9.9948e-01} & -1.6792e-04 & -3.3130e-04 & $y$ & \Blue{4.5274e+00} & \Green{7.2215e+00} & 1.4872e-07\\
    $\mu$   & -2.5933e-04 & 5.7727e-04 & 4.2798e-04 & $z$ & 5.4791e-07 & -3.4551e-06 & \Blue{-2.6812e+00}\\
    $x^2$   & -3.8502e-04 & 7.2785e-04 & -2.8368e-04 & $x^2$ & -1.7573e-06 & 1.7074e-06 & 1.4821e-02\\
    $xy$    & 1.6046e-03 & -4.7278e-05 & 1.6167e-03 & $xy$ & -1.4202e-06 & 1.6468e-06 & \Red{9.6046e-01}\\
    $x\mu$  & \Red{9.9672e-01} & -3.5927e-04 & -5.1250e-04 & $xz$ & -9.4857e-01 & \Blue{4.6996e-01} & -3.5173e-06\\
    $y^2$   & 3.8864e-04 & 3.1930e-04 & -2.0894e-04 & $y^2$ & -1.5292e-06 & 8.3760e-06 & 1.6890e-02\\
    $y\mu$  & -4.5870e-04 & \Red{9.9483e-01} & -8.0084e-04 & $yz$ & 4.5105e-01 & -6.7203e-01 & -8.0195e-06\\
    $\mu^2$ & 2.6971e-04 & 5.6016e-04 & 2.3888e-04 & $z^2$& -8.3810e-05 & 1.1768e-04 & 2.0347e-03\\
    $x^3$   & \Green{-9.9727e-01} & 1.3899e-05 & -1.2669e-03 & $x^3$& -5.9848e-02 & 9.4702e-02 & -8.1650e-07\\
    $x^2y$  & -2.1682e-03 & \Green{-9.9460e-01} & -5.3697e-04 & $x^2y$& 6.0315e-02& -8.9793e-02 & -4.1083e-06\\
    $x^2\mu$& -2.4824e-04 & 5.5955e-04 & -2.7237e-04 & $x^2z$& -5.6009e-06 & 5.7127e-06 & -1.1296e-04\\
    $xy^2$  & \Yellow{-9.9607e-01} & -1.2029e-03 & -4.7030e-04 & $xy^2$& -1.9930e-02 & 2.6786e-02 & 1.4347e-06\\
    $xy\mu$ & 5.4106e-04 & 8.6669e-04& -8.0186e-04 & $xyz$& 1.1614e-05 & -1.3253e-05 & 8.0456e-04\\
    $x\mu^2$& -9.1186e-04 & 1.8879e-06 & -1.3931e-04 & $xz^2$& 2.1789e-02 & -3.3973e-02 & -3.2658e-07\\
    $y^3$   & 7.2948e-05 & \Yellow{-9.9499e-01} & -1.5385e-04 & $y^3$& 2.8419e-03 & -3.5977e-03 & -1.2999e-07\\
    $y^2\mu$& 3.1669e-04 & -2.1393e-04 & 1.0944e-03 & $y^2z$& 4.1031e-07 & -4.9989e-07 & -3.6131e-04\\
    $y\mu^2$& -4.0919e-04 & -5.7416e-04 & 7.1154e-04 & $yz^2$& -8.8740e-03 & 1.2924e-02 & 8.9630e-07\\
    $\mu^3$ & 3.7226e-04 & 1.8204e-04 & 9.8719e-05 & $z^3$& 2.7687e-06 & -3.6289e-06 & -6.3774e-05\\
    \hline
\end{tabular}}
\caption{[Hopf bifurcation and Lorenz] The coefficients learned by nSINDy via training without the magnitude-based pruning strategy.}
\label{tab:exp_set_appendix2}
\end{table}

\section{Comparison with MLP}
Figure depict time-instantaneous mean-squared errors measured between the ground truth trajectories and the trajectories computed from the learned dynamics functions, $\VelocityNN$. We consider the two models: the black-box neural ODE models and the proposed SINDy models. For the SINDy models, we use the same experimental settings considered in Experimental results Section. For the black-box neural ODE models, we consider feed-forward neural networks consisting of four layers with 100 neurons in each layer with the hyperbolic Tangent nonlinear activation (Tanh). For training the both models, we use the same optimizer, Adamax, and use the same initial learning rate 0.01 with the same decay strategy (i.e., exponential decay with the multiplicative factor 0.9987). Both models are trained over 500 epochs and 2000 epochs, respectively, for the first four benchmark problems and the last benchmark problem, respectively. 
\begin{figure}[bt]
    \centering
     \includegraphics[scale=.55]{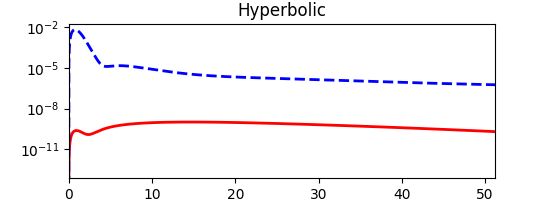}\\
    \includegraphics[scale=.55]{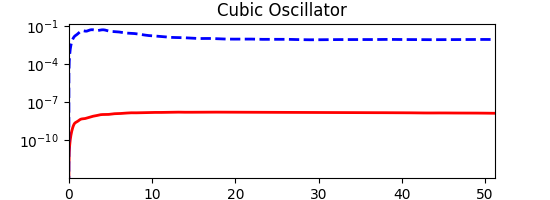}\\
    \includegraphics[scale=.55]{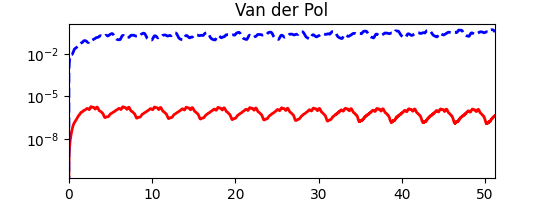}\\
    \includegraphics[scale=.55]{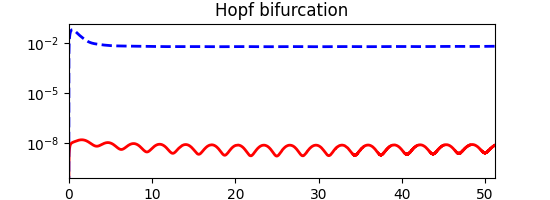}\\
    \includegraphics[scale=.55]{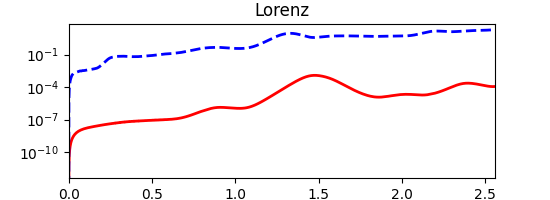}
    \caption{[nSINDy experiments] Time-instantaneous mean-squared errors. Two model are compared: the black-box neural ODE models (blue dashed lines) and the proposed neural SINDy model (red solid lines). }
    \label{fig:appdix_set1}
\end{figure}

\section{Port-Hamiltonian structure preservation}
Here, we consider the dictionary-based parameterization for the  port-Hamiltonian systems. We follow the port-Hamilonian neural network formulation presented in \cite{desai2021port}, which is written as:
\begin{equation}
    \begin{bmatrix}
    \dv{q}{\TimeSymb}\\[8pt]
    \dv{p}{\TimeSymb}
    \end{bmatrix} 
    = \begin{bmatrix*}[r]
    -\pdv{\Hamiltonian_{\NNParams}}{p}\\[8pt]
    \pdv{\Hamiltonian_{\NNParams}}{q}
    \end{bmatrix*}
    + \begin{bmatrix}
    0 \\[8pt]
    N \pdv{\Hamiltonian_{\NNParams}}{p}
    \end{bmatrix}
    + \begin{bmatrix}
    0 \\[8pt]
    F(t)
    \end{bmatrix},
\end{equation}
where $\Hamiltonian_{\NNParams}$ denotes the parameterized Hamiltonian function, $N$ denotes the nonzero coefficient for a damping term, and $F(t)$ denotes a time-dependent forcing. We consider the dictionary-type parameterization: 
\begin{equation}
    \Hamiltonian_{\NNParams} = (\DictionaryVec(q,p)\Transpose \CoeffsMat)\Transpose,
\end{equation}
set $N$ to be a trainable coefficient, and assume that $F(t)$ is known. 

As in \cite{desai2021port}, we consider the Duffing equation, where the Hamiltonian function is defined as
\begin{equation}
    \Hamiltonian(q,p) = \frac{p^2}{2m} + \alpha \frac{q^2}{2} + \beta \frac{q^4}{4}.
\end{equation} 
With the Hamiltonian, the Duffing equation considering the damping term and the forcing can be written as:
\begin{equation}
    \ddot{q} = -\alpha q - \beta q^3 + \gamma \sin(\omega t) -\delta \dot{q}.
\end{equation}

In the following experiment, we consider $\PolySet(2,4) = \{1, q, p, q^2, qp, p^2, q^3, q^2p qp^2, p^3, q^4, q^3p, q^2p^2, qp^3, p^4 \}$ for parameterizing the Hamiltonian function. 

\paragraph{Nonchaotic case:} 
$\alpha = -1$, $\beta = 1$, $\gamma=0.3$, $\delta =0.3$, and $\omega=1.2$. In the experiment, we assume that we know the values of $\gamma$ and $\omega$ and identify the Hamiltonian function. The identified Hamiltonian function is $\Hamiltonian_{\NNParams}(q,p) = 0.50015 p^2 -0.50004 q^2 + 0.25010 q^4$ and the identified value of $\delta = 0.3002$.

\paragraph{Chaotic case} 
$\alpha = -1$, $\beta = 1$, $\gamma=0.1$, $\delta =0.39$, and $\omega=1.4$. The identified Hamiltonian function is $\Hamiltonian_{\NNParams}(q,p) = 0.50007 p^2  -0.50053 q^2 + 0.25017 q^4$ and the identified value of $\delta = 0.1001$.

\begin{figure}[!tb]
    \centering
     \includegraphics[scale=.55]{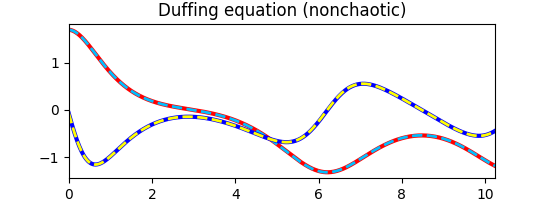}
    \includegraphics[scale=.55]{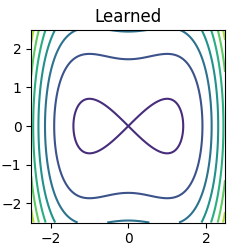}
    \includegraphics[scale=.55]{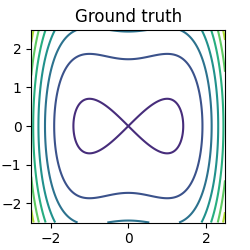}
    \caption{[Duffing equation] Left:  Examples of reference trajectories (solid lines) and computed trajectories (dashed lines) from learned dynamics. Right: the learend and the ground truth Hamiltonian function evaluated on the uniform grid of $(q,p)$.}
    \label{fig:portHNN}
\end{figure}
\end{document}